\documentclass[11pt,letterpaper]{article}
\usepackage{cogsys}

\usepackage{hyperref}
\usepackage[T1]{fontenc}
\usepackage{times}
\usepackage{latexsym}
\usepackage[pdftex]{graphicx} 
\usepackage{tabularx}
\usepackage[inline]{enumitem}

\usepackage{natbib}
\cogsysheading{11}{2021}{1--20}{9/2021}{11/2021}

\ShortHeadings{Scales and Hedges in a Logic with Analogous Semantics}
              {H.R.\ Schmidtke and S. Coelho}

\usepackage{amssymb}
\usepackage{amsmath}
\usepackage{latexsym}

\newcommand{\angled}[1]{\langle#1\rangle}

\newcommand{\currchap}{0}

\newcommand{\figref}[1]{Figure~\ref{fig:#1}}
\newcommand{\Figref}[1]{Figure~\ref{fig:#1}}
\newcommand{\tabref}[1]{Table~\ref{tab:#1}}
\newcommand{\Tabref}[1]{Table~\ref{tab:#1}}

\newcommand{\secref}[1]{Section~\ref{sec:\currchap.#1}}

\newcommand{\Secref}[1]{Section~\ref{sec:\currchap.#1}}

\newcommand{\sign}[1]{\csname#1\endcsname}
\newcounter{defc}

\newcounter{thc}

\newcounter{axc}

\newcounter{ruc}

\newcommand{\figlab}[1]{\label{fig:#1}}
\newcommand{\tablab}[1]{\label{tab:#1}}

\newcommand{\seclab}[1]{\label{sec:\currchap.#1}}

\newcommand{\defarrow}{\stackrel{\scriptscriptstyle{\mathit{def}}}
									{\Leftrightarrow}}

\newcommand{\limp}{\rightarrow}
\newcommand{\lbiimp}{\leftrightarrow}
\newcommand{\partof}{\ensuremath{\sqsubseteq}}
\newcommand{\ppartof}{\ensuremath{\sqsubset}}
\newcommand{\ipartof}{\ensuremath{\sqsupseteq}}
\newcommand{\ovl}{\bigcirc}

\newcommand{\opartof}{\mathbin{\mathaccent\cdot\sqsubseteq}}
\newcommand{\sqand}{\sqcap}
\newcommand{\sqor}{\sqcup}
\newcommand{\comp}{\ensuremath{{\sim}}}

\begin{document}

\title{Scales and Hedges in a Logic with Analogous Semantics}

\author{H.R. Schmidtke}{schmidtke@acm.org} 
\address{Independent / Postfach 11 01 29 Schwerin, 19001 Germany}
\author{S. Coelho}{arascoelho@gmail.com} 
\address{Faculty of Medicine, University of Lisbon / Av. Prof. Egas Moniz, 1649-028 Lisbon, Portugal}

\vskip 0.2in

\begin{abstract}
Logics with analogous semantics, such as Fuzzy Logic, have a number of explanatory and application advantages, the most well-known being the ability to help experts develop control systems. From a cognitive systems perspective, such languages also have the advantage of being grounded in perception. For social decision making in humans, it is vital that logical conclusions about others (cognitive empathy) are grounded in empathic emotion (affective empathy). Classical Fuzzy Logic, however, has several disadvantages: it is not obvious how complex formulae, e.g., the description of events in a text, can be (a) formed, (b) grounded, and (c) used in logical reasoning. The two-layered Context Logic (CL) was designed to address these issue. Formally based on a lattice semantics, like classical Fuzzy Logic, CL also features an analogous semantics for complex fomulae. With the Activation Bit Vector Machine (ABVM), it has a simple and classical logical reasoning mechanism with an inherent imagery process based on the Vector Symbolic Architecture (VSA) model of distributed neuronal processing. This paper adds to the existing theory how scales, as necessary for adjective and verb semantics can be handled by the system.
\end{abstract}

\section{Introduction}\seclab{intro}
Logics with analogous semantics have a number of advantages over conventional set-theoretical semantics. A primary advantage is that a sematics based on vector spaces can be related to a representation of physical reality in terms of vector spaces \citep{Gaerdenfors.2000}. In a literal sense, a spatial semantics over a coordinate space yields an image of the world. The \emph{symbol grounding} problem \citep{Harnad.2003} -- vital for understanding higher cognitive functions --, becomes easier with such a -- multidimensional -- image semantics, since an image can be compared to an image generated by means of sensory data \citep{Schm2020TxtMapAlg}. 

While questions regarding higher cognitive functions, such as \emph{symbol grounding} or our sense of \emph{meaning}, used to be philosophical rather than practical questions of immediate concern, advances in autonomous vehicles have brought the lack of meaning in current computer systems, including AI systems, into the public debate. 
Context Logic 
\citep[CL][]{Schm2020Multi,Schm2021Reaso,Schm2021Towar,Schm2020TxtMapAlg,Schmidtke2019LogRot,schmidtke2018,Schmidtke2018Canvas,schmidtke2016granular,Schmidtke.2014,Schmidtke.2013,schmidtke2012contextual,schmidtke2011distributed,SchmBeig2010,SchmWoo.2009,SchmHongWoo.2008,HongSchWoo.2007} as a cognitively motivated logic with analogous semantics is a promising candidate for addressing this shortcoming. 
Together with its biologically inspired reasoner/imager -- the \emph{Activation Bit Vector Machine} (ABVM) it becomes feasible to deliver a concept of meaning that more closely resembles a cognitively plausible inner world model semantics. 
This paper adds a discussion on scales to the theory and inllustrates that CL and the ABVM can be applied to ethical problems, such as the popular Trolley Problem discussed in the literature on ethics for autonomous vehicles \citep[e.g.,][]{SutfGast2017Using}.

Three key aspects are important for applying the theory to the ethical domain: first, how temporal notions, as used in the literature on decision making systems and planning, can be defined in the formalism; second, how scales, including ethical value scales, can be derived; and third, how the research can be related to human social reasoning. The first point connects the theory to existing domain theories on ethical reasoning and illustrates how to add analogous semantics to knowledge-based systems based on these domain theories. The second point is relevant to illustrate that the analogous semantics are analogous and cognitively adequate in specific key aspects, such as context dependency and the ability to apply linguistic means that allow the derivation of further values on a scale. The third point allows us to suggest the theory as a fine-grained tool to, on the one hand, better understand human pathologies and, on the other hand, better analyze and predict social and jurisdictional aspects of different types of AI systems. 

Conventional extensional set-theoretical semantics of first order logic (FOL) conceived as a formalization of scientific language and thought following \cite{frege1879begriffsschrift} 
was a crucial advancement for scientific inquiry and progress \citep{whitehead1912principia}: it provides mathematical rigidity to notions of scientific reasoning and a binary decision, true or false, sufficient for many purposes, such as when reviewing a scientific paper. 
From the perspective of cognitive science and commonsense reasoning, however, conventional set-theoretical semantics are unsatisfactory. 
Commonsense suggests that, e.g., a fictional novel creates a dynamical model within the mind of the reader, an inherent meaning. 
This commonsense notion of meaning is of crucial relevance as we move from lab experiments to autonomous robotic systems performing what we would call with a human actor unethical actions. We are at a crossroads: do we either expand our theory of semantics or do we re-educate our ethical sense to accept what is technically feasible under a previous paradigm?

An early example of a family of logics with atom-level analogous semantics are Fuzzy Logics \citep{zadeh1975fuzzy,zadeh1988fuzzy}. The original conception of Fuzzy Logic addressed the issue by assigning analogous values to statements, such as ``this apple is red,'' e.g., by calculating the average percentage of red in the RGB value of an area called \emph{the apple} in a picture. The result is a value in the set $[0,1]$ and by definition directly corresponds to reality. Accordingly, it can be used to control systems in a quasi-linguistic manner \citep{passino1998fuzzy}. There are three primary disadvantages with the original conception if we interpret it cognitively for charactizing the semantics of adjectives. First, adjectives like ``red'' are \emph{context dependent}. A red wine and a red car would be imagined as having a different type of red. Likewise, we change our conception of what size is referred to as ``tall'' when we hear the speaker is talking about a child or a woman. Second, linguistic research on the meaning of adjectives suggests that the unary, predicative use with the positive in ``\emph{x} is tall'' is semantically derived from the binary, comparative use in ``\emph{x} is taller than \emph{y}'' \citep{bierwisch1989semantics}. Most importantly however, the analogous semantics of conventional Fuzzy Logic regards only the level of atoms. The semantics for compound sentences discussed in the literature focusses mostly on derivatives of classical $[0,1]$-valued semantics, that is: the meaning of a sentence is a value in $[0,1]$, which is still far away from a dynamical model, an analogous sentence semantics. The class of semantics for Fuzzy Logic is larger, however, as shown by \cite{hajek1998metamathematics}. All lattice structures, of which those based on the totally ordered $[0,1]$ are only a subset, have the basic properties. However, it was less clear what practical use within the Fuzzy Logic framework there could be for arbitrary lattices, e.g., what the lattice spanned by the subsets of a finite set with the subset-relation could contribute.

But lattice structures have been the fundament of another, earlier research thrust to avoid the issues of a set-theoretical foundation of logic: \emph{mereology}. Le\'sniewski's Protothetic \citep{SrzeSta.2012} aimed at building the fundamentals of logic on binary relations \emph{part of} to fill the role of the set-theoretical subset-relation. However, the direct application of mereology to provide analogous semantics to complex statements, e.g., in the form of images resembling Venn-diagrams \citep{Sowa2008Conce} only generates visualizations of the set-theoretical semantics. The images may be of pedagogical use to learn set-theoretical semantics, but they do not resemble the represented world in the same strong sense in which Fuzzy Logic allows the reconstruction of the color of the red apple in our initial example. They also cannot help us to understand, for instance, the vivid mental mental model and strong feeling of disgust evoked when reading a description of how a driver was mutilated by his autonomous car.

The CL family of languages, like mereology, leverages a partial order ($\partof$, which can be read as an abstract notion of ``part of'' as in mereology or as ``sub-context'')  and lattice structures at its core, but the image generation mechanism is based on an activation measurement process that has both neuronal and logical roots. The vizualization capabilities arise directly from its semantics and allow, e.g., the reconstruction of complex layouts from spatial language in a uniquely defined manner. CL has been developed since 2006 as a logical language at the boundary between perception and reasoning. A wealth of results have been discovered in the past 15 years \citep{Schm2020Multi,Schm2021Reaso,Schm2021Towar,Schm2020TxtMapAlg,Schmidtke2019LogRot,schmidtke2018,Schmidtke2018Canvas,schmidtke2016granular,Schmidtke.2014,Schmidtke.2013,schmidtke2012contextual,schmidtke2011distributed,SchmBeig2010,SchmWoo.2009,SchmHongWoo.2008,HongSchWoo.2007}. While we do not assume any familiarity with CL in this paper and will, for purposes of completeness, outline the language, its semantics, reasoner, imager, and so on, it is obviously beyond the scope of one paper to discuss all details and prove all the theorems again. The interested reader is referred to the respective publications.  

Most recently, it has been confirmed that CL can be understood as a member of the Fuzzy Logic family, and that it can even be further extended into a Fuzzy Context Logic (FCL) that unites the respective perspectives \citep{Schm2021Towar}. However, many interesting questions remain open in this regard, as the connection between this new, more powerful logic and CL's analogous semantics remains to be studied. If FCL has an analogous semantics for complex formulae like CL, conventional Fuzzy Logic would have one as well.

\paragraph{Structure of the Article}
The paper starts with a short introduction of the background theories of Context Logic and the ABVM (\secref{bck}). 
We then explicate how the theory can be applied in the ethical domain (\secref{xpl}), 
before we move to discussing the importance of ethical grounding in \secref{dsc} by comparing two social disorders, one associated with unethical behavior (psychopathy) due to impaired grounding of ethical concepts and one not associated with unethical behavior (autism) where grounding is intact. \Secref{cnc} concludes the paper summarily emphasizing the importance of grounding.

\section{Background}\seclab{bck}
This section provides a brief introduction to Context Logic \secref{cll} and the Activation Bit Vector Machine (ABVM) reasoner and imager \secref{img}. Given the limited space, the reader in doubt about a particular aspect is referred to the respective previous publications \citep{Schm2020Multi,Schm2021Reaso,Schm2021Towar,Schm2020TxtMapAlg,Schmidtke2019LogRot,schmidtke2018,Schmidtke2018Canvas,schmidtke2016granular,Schmidtke.2014,Schmidtke.2013,schmidtke2012contextual,schmidtke2011distributed,SchmBeig2010,SchmWoo.2009,SchmHongWoo.2008,HongSchWoo.2007}.

\subsection{Context Logic}\seclab{cll}
Formally, CL is a two-layered logic. 
\begin{enumerate}[noitemsep]
\item \emph{Context terms} $\mathcal{T}_C$ are defined over a set of variables $\mathcal{V}_C$: 
\begin{enumerate}[label=TY\arabic*,noitemsep]
 \item Any context variable $v \in \mathcal{V}_C$ and the special symbols $\top$ and $\bot$ are atomic context terms.
 \item If $c$ is a context term, then its complement $(\comp c)$ is a context term. 
 \item If $c$ and $d$ are context terms then the intersection $(c \sqand d)$ and sum $(c \sqor d)$ are context terms.
\end{enumerate}
\item \emph{Context formulae} $\mathcal{F}_C$ are defined as follows:
\begin{enumerate}[label=FY\arabic*,noitemsep]
 \item If $c$ and $d$ are context terms then $[c \partof d]$ is an atomic context formula.\label{atom}
  \item If $\phi$ is a context formula, then $(\lnot \phi)$ is a context formula.\label{lnot}
 \item If $\phi$ and $\psi$ are context formulae then $(\phi \land \psi)$, $(\phi \lor \psi)$, $(\phi \limp \psi)$, and $(\phi \lbiimp \psi)$ are context formulae.\label{lops}
 \item If $x \in \mathcal{V}_C$ is a variable and $\phi$ is a formula, then $\forall x: \phi$ and $\exists x: \phi$ are context formulae.\label{qunt}
\end{enumerate}
\end{enumerate}
In the following, we leave out brackets as far as possible applying the following precedence: $\comp,\sqand,\sqor$ for term operators and $\lnot,\land,\lor, \limp, \lbiimp$ for formula operators. The scope of quantifiers is to be read as maximal, i.e., until the first bracket closes that was opened before the quantifier, or until the end of the formula. Brackets around atomic formulae are used for easier visual separation between term layer and formula layer. 

The language's syntax gives rise to a hierarchy of sub-languages: $CLA \subset CL0 \subset CL1$, with the subset relation holding both syntactically as well as semantically. 
The CLA fragment (atomic CL) allows only $\land$, CL0 (propositional CL) allows any construction without quantifiers (\ref{lnot}, \ref{lops}), CL1 (first order CL) adds quantifiers (\ref{qunt}). We sketch the semantic properties axiomatically \citep{Schm2021Towar,schmidtke2012contextual} 
and characterize $\partof$ as a partial order, i.e., as reflexive \eqref{refl}, antisymmetric \eqref{antisym}, and transitive \eqref{trans} \citep[cf.\ ][ for a more detailed treatment]{Schm2021Towar}:
\begin{gather}
[\alpha \partof \alpha] \label{refl}\\
 \begin{aligned}[]
 [\alpha \partof \beta] \land [\beta \partof \alpha]  \limp  
 \forall \xi&: ([\alpha \partof \xi] \limp [\beta \partof \xi])  
\land ([\xi \partof \alpha] \limp [\xi \partof \beta]) \label{antisym}
 \end{aligned}\\
[\alpha \partof \beta] \land [\beta \partof \gamma]  \limp  [\alpha \partof \gamma]\label{trans}
\end{gather}
where $\alpha, \beta, \gamma, \xi$ are schema variables. 
The transitivity axiom \eqref{trans} allows us to move any complex context term, represented with an arbitrary unused context variable $x$, to the right-hand side. If a context $\beta$ is subcontext of a context $\gamma$, then a context $x$, subcontext of $\beta$ must be subcontext of $\gamma$:
\begin{gather}
[\beta  \partof \gamma] \limp ([x \partof \beta] \limp [x \partof \gamma]) \label{partof}
\end{gather}
This characterization allows us to move any complex context term to the right hand side. It is then sufficient to characterize the context term operators only with respect to their occurrence on the right hand side.
\footnote{The formulation is advantageous as it illustrates how a decidable conventional reasoner can operate over context logic formulae. This reasoning procedure can be considered as cognitively motivated in so far as it procedes by zooming into a context (left hand side of $\partof$), moving it to the right hand side \eqref{partof}, and then addressing its parts. We see that the process will terminate after finitely many steps as the number of term operators is decremented with each step \citep[see][for a detailed proof of decidability]{SchmHongWoo.2008}.}
\begin{gather}
[\top  \partof \alpha \sqand \beta] \lbiimp [\top  \partof  \alpha] \land [\top  \partof \beta] \label{sqand}\\
 \begin{aligned}[]
[\top  &\partof \alpha \sqor \beta] \lbiimp 
\forall \xi: \exists \chi: [\chi \partof \xi] \land ([\chi  \partof  \alpha] \lor [\chi  \partof \beta])
 \end{aligned}\label{sqor}\\
[\top  \partof \comp \alpha] \lbiimp [\alpha \partof  \bot] \label{comp}
\end{gather}
The formula \eqref{sqand} states that if the current context $\top$ is a subcontext of the intersection of $\alpha$ and $\beta$, then it is both a subcontext of $\alpha$ and of $\beta$. In contrast, if the current context $\top$ is a subcontext of the sum of $\alpha$ and $\beta$ \eqref{sqor} it need not be completely in either $\alpha$ or $\beta$ -- e.g., Russia is in Eurasia, the sum of Europe and Asia, but not in Europe or in Asia. Thus \eqref{sqor} can only demand that any subcontext has a subcontext in $\alpha$ or $\beta$ -- e.g., all parts of Russia have parts that are in Europe or in Asia. If $\top$ is in the complement of $\alpha$, $\alpha$ is outside of the current context \eqref{comp}.

The converse ($\ipartof$), equality ($=$), overlap  ($\ovl$), and a non-empty variant ($\opartof$) can be defined as:
\begin{align}
[ \alpha \ipartof \beta ] &\defarrow [\beta \partof \alpha ]  &
[ \alpha = \beta ] &\defarrow [ \alpha \partof \beta ] \land [\beta \partof \alpha ] 
\label{cladefs} \\
[ \alpha \ovl \beta ] &\defarrow \lnot [ \alpha \sqand \beta \partof \bot]  &
[ \alpha \opartof \beta ] &\defarrow [ \alpha \ovl \beta ] \land [ \alpha \partof \beta ]
\label{cl0defs}
\end{align}
The converse ($\ipartof$) relation is useful to specify opposites, such as \emph{north} and \emph{south} \citep{Schm2020TxtMapAlg}. Two contexts $\alpha$ and $\beta$ are equal ($=$) iff they are mutually subcontexts of each other. Overlap ($\ovl$) between $\alpha$ and $\beta$ means, that the intersection of $\alpha$ and $\beta$ is not empty. The non-empty variant of $\partof$ ($\opartof$) is defined by $\alpha$ being part of $\beta$ and overlapping it. 
The definitions of $\ipartof$ and $=$ \eqref{cladefs} are still in CLA. But the use of $\lnot$ for $\ovl$ and thus $\opartof$ \eqref{cl0defs} requires at least CL0, i.e., a CL0 reasoner \citep{Schm2020Multi}. 
It is a basic property of lattice theory that using the conjunctive term operator $\sqand$, further p.o.\ relations with corresponding variants can be constructed from the sublattices under elements. 
E.g., \emph{spatial-part-of} or \emph{north-of} can be modeled as p.o.\ relations based on contexts $P$ and $N$ \citep{schmidtke2011distributed}. For any $\xi, \alpha , \beta , \gamma$,
reflexivity \eqref{x.refl}, antisymmetry \eqref{x.antisym} and transitivity \eqref{x.trans} properties of a derived relation represented by a context $x$ follow immediately from the p.o.\ properties of $\partof$ together with the properties of $\sqand$ \citep{schmidtke2012contextual}. 
\begin{gather}
[ \alpha \sqand x \partof \alpha ] \label{x.refl}\\
[ \alpha \sqand x \partof \beta ] \land [\beta \sqand x \partof \alpha ]  \limp  
[\beta \sqand x = \alpha   \sqand x] \label{x.antisym}
 \\
 [\alpha \sqand  x  \partof \beta ] \land [\beta \sqand  x  \partof \gamma]  \limp  
 [\alpha \sqand  x  \partof \gamma] \label{x.trans}
\end{gather}
The proofs follow by \eqref{sqand} and \eqref{partof} via \citep[cf.][for details]{schmidtke2012contextual}:
\begin{gather}
[\alpha \sqand  x  \partof \beta ] \lbiimp [\alpha \sqand  x  \partof \beta \sqand  x ]\label{bidir}
\end{gather}
The intersection of $\alpha$ and $x$ is subcontext of $\beta$ if and only if the intersection of $\alpha$ and $x$ is subcontext of the intersection of $\beta$ and $x$ \eqref{bidir}. This holds because any subcontext that is in the intersection of $\alpha$ and $x$ is trivially also in $x$.
We can define a \emph{contextualization} syntax as a shorthand. 
\begin{gather}
 x : [\alpha \partof \beta ] \defarrow  [\alpha \sqand  x  \partof \beta ]. \label{contsyn}
\end{gather}
With respect to $x$, $\alpha$ is subcontext of $\beta$, iff the subcontext of $\alpha$ in $x$ is subcontext of $\beta$ \eqref{contsyn}. For instance, a (physical) conference is \emph{spatially} a subcontext of a certain convention center and \emph{temporally} in a certain month. But containment relations are not the only partial order relations, e.g.: the resulting trajectory of a billiard ball is caused by the angle, speed, and spin at which it hit another ball, and this, in turn, was caused by the angle, speed, and position at which it was hit by the billiard cue. The reflexive variant of causation, whatever other properties causation may have, is antisymmetric and transitive, i.e., belongs to the class of a partial orders.

Together with the existential quantitfier $\exists$, CL1 allows any other relation to be constructed including non-p.o.\ relations, such as the instance-of relation between an object $o$ and a class $c$:
\begin{gather}
\exists e: [e \partof \mathit{isi}] \land [o \sqand e \opartof a_1] \land [c \sqand e \opartof a_2].
\end{gather}
We can express that $o$ is an \emph{instance of} $c$ in CL by saying that there is a subcontext $e$ of \emph{isi} (intuitively, the edge $e$ of the graph \emph{isi}) of so that $o$ overlaps $e$ in $a_1$ (first arguments or ends of edges) and $c$ overlaps $e$ in $a_2$ (second arguments or tips of edges). 
Forrmally, this construction is a \emph{tuple generator}, with which arbitrary relations can be constructed. 

We can write both types of relational constructions as well as other derived relations in the conventional manner of standard FOL by using Context Logic schemata, e.g.:
\begin{gather}
 \mathit{starts}(\alpha,\beta) \defarrow [\alpha \sqand c \partof \beta] \land [\alpha \sqand t \partof \beta]\\
\begin{aligned}
 \mathit{isi}(\alpha,\beta) \defarrow \exists e&: [e \partof isi]
 \land [\alpha \sqand e \opartof a_1] \land [\beta \sqand e \opartof a_2].
\end{aligned}
\end{gather}
We can use square brackets to indicate a relation is a partial order, thus further shortening \eqref{contsyn}:
\begin{gather}
x[\alpha, \beta] \defarrow  [\alpha \sqand x \partof \beta]. \label{pepo}
\end{gather}
We thus have derived the conventional syntax of \emph{predicate expressions} demonstrating that these can be considered internally complex constructions. We gain the advantage of reducing the number of axioms required for basic and compound transitive relations, such as the above \emph{starts} (see below) and have shown that CL does not replace conventional predicate logic but adds a way to further analyze and better understand its atomic formulae. 

\subsection{Analogous Semantics of Context Logics}\seclab{img}
If we compare classical propositional logic with FOL, a key distinction of FOL is its semantics' dependence on external set-theoretically specified structures. 
Truth in propositional logic, in contrast, does not depend on additional structures. 
A formula with $n$ variables has exactly $2^n$ possible assignments. 

In recent decades, a larger number of other decidable logics have been identified, especially in the family of Modal Logics and Description Logics. A particularly interesting discovery regards the status of partial order relations yielding decidable logics as well \citep{KierTend2018Finit}. P.o.s are both historically relevant, at the heart of Intuitionistic Logic \citep{ChagZakh.1997} and Mereology \citep{SrzeSta.2012}  
as well as the core of modern type and class based object-oriented reasoning systems \citep{BracLeve.2004}. Propositional Context Logic (CL0) is the decidable fragment of Context Logic \citep{Schm2021Towar} featuring $\partof$ as the single partial order relation \cite{Schm2021Towar}. Without CL0's negation, like intuitionistic negation a more powerful  tool than propositional logic negation, the more basic CLA is even equivalent to propositional logic. This allows us to bring CLA reasoning down to a neuronal bit-level reasoning procedure, where we can also connect it to high-level theories of neuronal computation. 

Vector Symbolic Algebras \citep[VSA, ][]{kanerva2009hyperdimensional} are a cognitively plausible yet computationally parsimonious model of  neuronal network operation \citep{gayler2006vector}. Apart from computational advantages, such as low energy requirements, they are also a logically interesting format. The basis of a traditional VSA system are long binary random vectors -- although other variants exist. Traditional VSA systems focus, on the one hand, on modelling cognitively plausible associative learning, retrieval, and reasoning mechanisms, and, on the other hand, on the replication of symbolic computing mechanisms, such as pointers and tuples. 
The Activation Bit Vector Machine \citep[ABVM,][]{Schm2020Multi,Schm2021Reaso,Schm2020TxtMapAlg,schmidtke2018,Schmidtke2018Canvas} leverages the same infrastructure, but only uses logical operations: as the operation retaining similarity, we use the bit-wise \emph{or} ($|$), which behaves in the same way as the averaging sum on sparse vectors.

\paragraph{Focus and Filter}
If we interpret a VSA logically, we can note that it behaves similarly to a truth table, the standard semantics of Propositional Logic. If a binary vector $a$ encodes one formula's possible models and a vector $b$ encodes another formula's models and the vector operation $a \& b$ does not have any 1s, at all, then $a$ and $b$ probably contradict each other. We cannot be certain, because we did not check systematically, but if both vectors were generated in a random manner and are large enough, then there is a high probability that we encountered each possible combination, as in a truth table. Moreover, the distinct rows of the truth table will have been encountered a proportional number of times in the randomized variant. More formally, we have a Monte Carlo Simulation of the truth table method and thus a linear probabilistic SAT and \#SAT reasoner. This system is very fast but like human working memory has tight limits, such as the human ${7\pm 2}$ items \citep{baddeley1994magical,miller1956magical}. 
If we operate with binary vectors of length $\kappa$, the parameter $\kappa$ corresponds to the number of Monte Carlo samples taken. With $n$ variables in a formula, we have approximately ${\kappa}/{2^n}$ 
occurrences of a combination. If we chose $\kappa < 2^n$, we necessarily could not cover all combinations. We can, however, already cover the human $n = 7\pm 2$ using the truth table method with $\kappa = 2^{7\pm 2}$, i.e., $32 \leq \kappa \leq 512$. With the random method, the probability to miss a combination of $n$ variables is 
$$\frac{2^n}{\kappa}\text{, i.e., for }n = 7\pm 2 \text{ and }\kappa = 10000: \frac{32}{10000} \leq p \leq \frac{512}{10000},$$
i.e., $0.0032 \leq p \leq 0.0512$ \citep{Schm2021Reaso}. 

CLA reasoning maps well to logical bit vector reasoning. With a spatial intuition for $\partof$ as part-of between regions, we can consider any position $i$ on vectors $a$ as a sampling point, which can either be inside ($a_i=1$) or outside of $a$ ($a_i=0$). 
The operation $\&$ provides us with a focus mechanism: $a_i \& b_i=1$ holds in those parts of $b$ that are also in $a$, so $a \& b$ is the $b$ part of $a$ or the $a$ part of $b$. Using negation (symbol: !), we can filter out parts we do not want to consider: $!a_i \& b_i=1$ holds for points in $b$ outside $a$. 
We can alternatively say $!a \& b$ subtracts $a$ from $b$ and obtain a filter mechanism. We can thus explain $\&$ and $!$ as arising from perceptual focus and foregrounding mechanisms. With $|$ leveraged for associative/similarity inference, all the binary operators ($!, \&$ and $|$) can be explained as being individually useful for fundamental perceptual and memory mechanisms. 

\paragraph{Partial Order}
A corresponding region-based partial order can be derived from Propositional Logic entailment. A formula $a$ entails a formula $b$, iff all assignments that make $a$ true also make $b$ true. In terms of sampling $a$ and $b$: if all points $i$ in $a$ ($a_i=1$) are also in $b$ ($b_i=1$) -- while points outside $a$ can be either in or outside $b$ -- then the p.o.\ relation holds ($[a \partof b]$). This can be phrased in terms of negation and conjunction as $(a_i\&!b_i)=0$ \emph{for all positions} $i$, or, alternatively, $!(a_i\&!b_i)=1$ \emph{for all positions} $i$. Note the difference between vector operators ($!, \&, |$) which we derive from binary operators and which applied to two vectors yield a vector, in contrast to the entailment relation which applied to two vectors yields either \emph{yes} or \emph{no}, i.e., a truth value. In terms of a classical logical semantics interpretation in VSAs, the only results we thus would be interested in would be whether $(a\&!b)$ yields the 0-vector, consisting only of 0s or not. We have a cognitively plausible linear probabilistic SAT reasoner for formulae with a small number of variables, such as $7\pm 2$. This is also the bridge between the term layer and the formula layer in CL.
We can describe different partial orders with the focus mechanism \citep{schmidtke2011distributed}. With the  above intuition about $\partof$, we also obtain arbitrary many other partial relations, such as \emph{spatial part of} or \emph{north of}. We can consider $p \& x$ to mean a focus on the positions $i$ where $x_i=p_i=1$. 

In order to encode an entire network of relations, we can use the operator $\&$ also for reading the logical operator $\land$, given that it it is closely related to $\sqand$ \eqref{sqand}. Note that, like for intuitionistic logic, we will need more for $\lnot$ and $\lor$ (CL0), but for CLA the bit vector logical operations are sufficient. For a knowledge base  (KB) $\phi=n:[a \partof b]\ \land\ n:[b \partof c]\land\ldots$, we obtain $\phi = !(a\&n\&!b)\&!(b\&n\&!c)\&\ldots$ as the vector encoding of the KB. We can query the KB, e.g., for $n:[a \partof c]$ by asking for the encoding of the query $q=!(a\&n\&!c)$, whether $!\phi |q$ is the 1-vector.

\paragraph{Relation to Conventional Set-Theoretical Semantics}  Summarizing we can see that the VSA semantics is a probabilistic variant of a classical set-theoretical semantics for CLA conceived of as a fragment of FOL with a single relation $\partof$. Note, that with infinitely long deterministically ordered vectors, we obtain a fragment of FOL with a unique, inherent semantics.

\paragraph{From Binary Vectors to Analogous Representations}\seclab{img.img}
We can do even more with the focus and filter mechanism. If we look at $\phi\&a$ we focus on all  information a formula $\phi$ has about the object $a$. If we look at $\phi\&n$ we focus on all  information $\phi$ has about the relation $n$. Moreover, $\phi\&n\&a$ allows us to focus on $a$ with respect to the $n$-relation. 
Resuming the previous example $\phi = !(a\&n\&!b)\&!(b\&n\&!c)\&\ldots$ and looking at vectors $\phi\&n\&a$, $\phi\&n\&b$, $\phi\&n\&c$, we see that with $\phi$ we remove -- i.e., set to 0 -- all positions $i$ where $\phi_i\&n_i\&a_i=1$ and $b_i=0$ and all portions where $\phi_i\&n_i\&b_i=1$ and $c_i=0$. In other words, all positions, where $\phi_i\&n_i\&c_i=1$ have $\phi_i\&n_i\&b_i=1$ and $\phi_i\&n_i\&a_i=1$, and all positions, where $\phi_i\&n_i\&b_i=1$ have $\phi_i\&n_i\&a_i=1$. This means that there can only be more (or equally many) 1s in $\phi\&n\&a$ than in $\phi\&n\&b$, which in turn has more 1s than $\phi\&n\&c$. Generalizing, $|\phi\&n\&x|$ the number of 1s in $\phi\&n\&x$ yields a numerical representation of the ordering regarding the north-aspect of $x$, a rough north-coordinate \cite[cf.][for larger examples and a more detailed discussion]{Schm2020Multi,Schm2020TxtMapAlg,Schmidtke2019LogRot,schmidtke2018}. 
Obviously, we can do the same for any number of relations. That is, we can take any two relations, e.g., north and east or size, but also, e.g., the health-dimension, its related emotional and ethical dimensions as well as the two temporal dimensions, and relate them to yield analogous values. These values can be fed backwards along the perceptual pathway to components lower in the pathway. The result of reasoning can be felt or seen, that is, \emph{imagined}. 

\section{Application in the Ethical Domain}\seclab{xpl}
In this section, we illustrate the process of domain modelling with CL, with particular focus on the ethical domain. As a concrete example, we use the popular Trolley problem \citep{foot1967}. We highlight a key step of modeling with CL, the disassembly of compound relations into grounded partial order core relations for the example of the temporal domain. 

\subsection{Natural Language Fragment}
\Tabref{lng} shows an excerpt of the small language fragment we use. We focus on simple predication and simple action sentences. To discuss a more realistic and concrete example with complex ethical and emotional dimensions we leveraged the following simple description \citep{Schm2020Multi} of the philosophical Trolley problem \citep{foot1967}:
\begin{quote}
 A trolley is moving down a track. If the agent pulls a lever, the trolley will move down a side track killing one person. If the agent does not pull the lever, the trolley will continue down the track killing five people.
\end{quote}
While we do not go into the details of the ethical and emotional processing of participants in experiments on this particular ethical problem \citep{Jafa2018Our-B}, we will return to the example and ethical and related emotional dimensions below (Sections \ref{sec:0.sch}-\ref{sec:0.cnc}). It is sufficient to note here, that pain is a core physical sensory experience related to loss of health (\tabref{lng}) and associated at a deep cognitive level, early in both development and evolution, with existential fear. The evaluation of something as causing pain, i.e., reducing health, or alleviating pain, i.e., felt as increasing health, is tied in social animals to the emotional/social dimensions of enmity and benevolence, respectively. 
The ethical dimensions (good and evil) are directly tied to the emotional/social dimensions. Ethical reasoning, however, requires the level of alternative scenarios (CL0), actions, and agency (CL1). At the core of ethical reasoning is reasoning about alternatives \citep{Schm2020Multi} as well as the awareness of an agent of their own position in the evaluation frame in dependence on their actions. An agent denying the existence of alternatives or their own agency in creating harm to another has cognitive issues at this higher level. However, an agent lacking the ability to tie harm to another to the emotional/social dimensions of enmity lacks a more fundamental capability found early in development and evolutionarily in many other mammals: \emph{affective empathy}, the capability to feel the pain of the other on a sensory dimension of compassion \citep{batson2009these}, in the case of a participant reading the Trolley Dilemma, the grounding of abstract symbols in emotion. 

\begin{table}[tb]
\caption{Excerpt from basic Vocabulary and Semantic Classes with Dimensional Meanings}
\begin{center}\small
\parbox{.45\linewidth}{
\begin{tabular}{c|c|c}
Type & Dimension & Comp. \\ \hline
Structural & - & is \\
Structural & - & are \\ \hline
PP (static) & north-south (+) & north \\
PP (static) & north-south (-) & south \\
PP (static) & east-west (+) & east \\
PP (static) & east-west (-) & west \\
PP (static) & size (+) & large \\
PP (static) & size (-) & small \\
PP (static) & left-right (-) & left \\
PP (static) & left-right (+) & right \\
PP (static) & left-right (*) & side (adv.) \\ \hline
PP (dynamic) & up-down (+) & up \\
PP (dynamic) & up-down (-) & down \\
PP (dynamic) & to-from (+) & to \\
PP (dynamic) & to-from (-) & from \\ \hline
Aspect & contains-during (+) & V-ing \\
Tense & before-after (+) & will V \\ 
\end{tabular}} \hfill
\parbox[t]{.45\linewidth}{
\begin{tabular}{c|c|c}
Type & Dimension & Comp. \\ \hline
Verb (i) & spatial (obj-ext.) & move \\ 
Verb (i) & spatial (obj-ext.) & run \\
Verb (i) & spatial (obj-ext.) & go \\
Verb (i) & spatial (obj-ext.) & continue \\
Verb (i) & health (+, neg to avg) & recover \\ \hline
Verb (t) & spatial (obj-ext.) & drive \\
Verb (t) & spatial (subj-arm) & pull \\
Verb (t) & health (+, neg to avg) & heal \\
Verb (t) & health (-) & harm \\
Verb (t) & health (min) & kill \\
Verb (t) & poss. space (subj, -) & give \\
Verb (t) & poss. space (subj, +) & receive \\ \hline
NP & - & a/the trolley \\
NP & - & a/the track \\
NP & - & an/the agent \\
NP & - & a/the lever \\
NP & - & one person \\
NP & - & five people \\
\end{tabular}}
\end{center}
\tablab{lng}
\end{table}
Whether a certain direction is considered primary (+) or converse (-) direction depends on the dimension. Measures of size, for instance, have a distinct positive direction, as size is a ratio attribute \citep{SuppZinn.1963} with a distinct zero and positive direction determined by physical reality. Other dimensions, such as the positive direction for left-right may be determined by characteristics of the speaker/listener such as handedness or direction of writing. 

\paragraph{Space}
Both spatial containment and spatial ordering dimensions can be covered by p.o.s. Moreover, the same absolute space can be conceptualized as having different dimensionality across contexts. The formalization of granular mereogeometry \citep{schmidtke2016granular} shows that both granularity and dimensionality can be conceived of as linearizations of the spatial containment p.o.\ illustrating the expressive power of the CL framework.  

Linguistically, the CLA level is sufficient for simple predicates, such as ``A is north of B.'' For eight of the nine static PP components in \tabref{lng} a simple p.o., $\angled{dim}^\sigma$, where $\sigma\in\{+,-\}$, is sufficient to capture the semantics:
\begin{equation}
  \angled{subj}  (is | are) \angled{pred^\sigma} \angled{obj} \mapsto 
  \angled{dim}^\sigma[\angled{subj}, \angled{obj}]
\end{equation}
E.g.: ``A is north\_of B.'' is mapped to $north\_of[A,B]$, i.e., by \eqref{pepo} to $north\_of: [A \partof B]$ which is equivalent to $[north\_of \sqand A \partof B]$ by  \eqref{contsyn}.
The dynamic PPs occur in conjunction with verbs. Action sentences require the more powerful CL1 framework, more specifically the ability to generate contexts that describe different situations or states of the world \citep{Schm2020Multi}. 

\paragraph{Time}
We can describe temporal succession and containment as specified by tense ($\tau$) and aspect ($\alpha$) of a verb in terms of two dimensions. 
By combining ordering relations \emph{before-after}, derived from causation ($c$), and \emph{contains-during} derived from local temporal interval containment ($t$) 
one can obtain a jointly exhaustive and pairwise disjoint relation system similar to that of \cite{allen1983maintaining}.\footnote{Note, that we do not define causation based on \emph{before-after}, but, just the opposite, define \emph{before-after} to be derived from the more immediate causation. The abstract time line is -- even one abstraction step further away -- formally a linear extension, i.e., a linear or total ordering derived from the relation shown here. Following the CL program, we would conjecture humans learned to, and children learn to, reccognize causation before developing \emph{before-after} and before developing a concept of linear time.} 
\Figref{intervals} shows the definitions (left) and illustrates how we can think about the difference between $c$ and $t$ (right). While $t$ provides interval containment (horizontal lines between boundary boxes), $c$ contributes the ordering properties (symbolized by a circle). We can think of $c$ as contributing a point-based representation of directedness, whereas $t$ focusses on temporal parts and overlap. Without further axioms, the two notions are independent. 
\begin{figure}[tb]
\begin{center}\parbox[c]{.6\linewidth}{\small
\begin{tabular}{>{$}c<{$}|>{$}c<{$}|>{$}c<{$}}
\text{relation name} & \text{causation ($c$)} & \text{containment ($t$)} \\ \hline
\mathit{core}(i,j) & i \sqand c = j \sqand c &  \\
\mathit{starts}(i,j) & i \sqand c \ppartof j & i \sqand t \ppartof j \\
\mathit{finishes}(i,j) & j \sqand c \ppartof i & \\ 
\hline
\mathit{icore}(i,j) & i \sqand c = j \sqand c & \\
\mathit{istarts}(i,j) & i \sqand c \ppartof j & j \sqand t \ppartof i \\
\mathit{ifinishes}(i,j) & j \sqand c \ppartof i &  \\ 
\hline
\mathit{qcore}(i,j) & i \sqand c = j \sqand c & \\
\mathit{qstarts}(i,j) & i \sqand c \ppartof j & j \sqand t = i \sqand t \\
\mathit{qfinishes}(i,j) & j \sqand c \ppartof i &  \\ 
\hline
\mathit{qovlc}(i,j) & i \sqand c = j \sqand c &  \\
\mathit{ovlc}(i,j) & i \sqand c \ppartof j & i \sqand t \circ j\\
\mathit{iovlc}(i,j) & j \sqand c \ppartof i &  \\ 
\hline
\mathit{meets}(i,j) & i \sqand c = j \sqand c &  \\
\mathit{before}(i,j) &i \sqand c \ppartof j & \lnot[i \sqand t \ovl j] \\
\mathit{after}(i,j) & j \sqand c \ppartof i & \\
\end{tabular}}\hfil
\parbox[c]{.27\linewidth}{
\includegraphics[width=\linewidth]{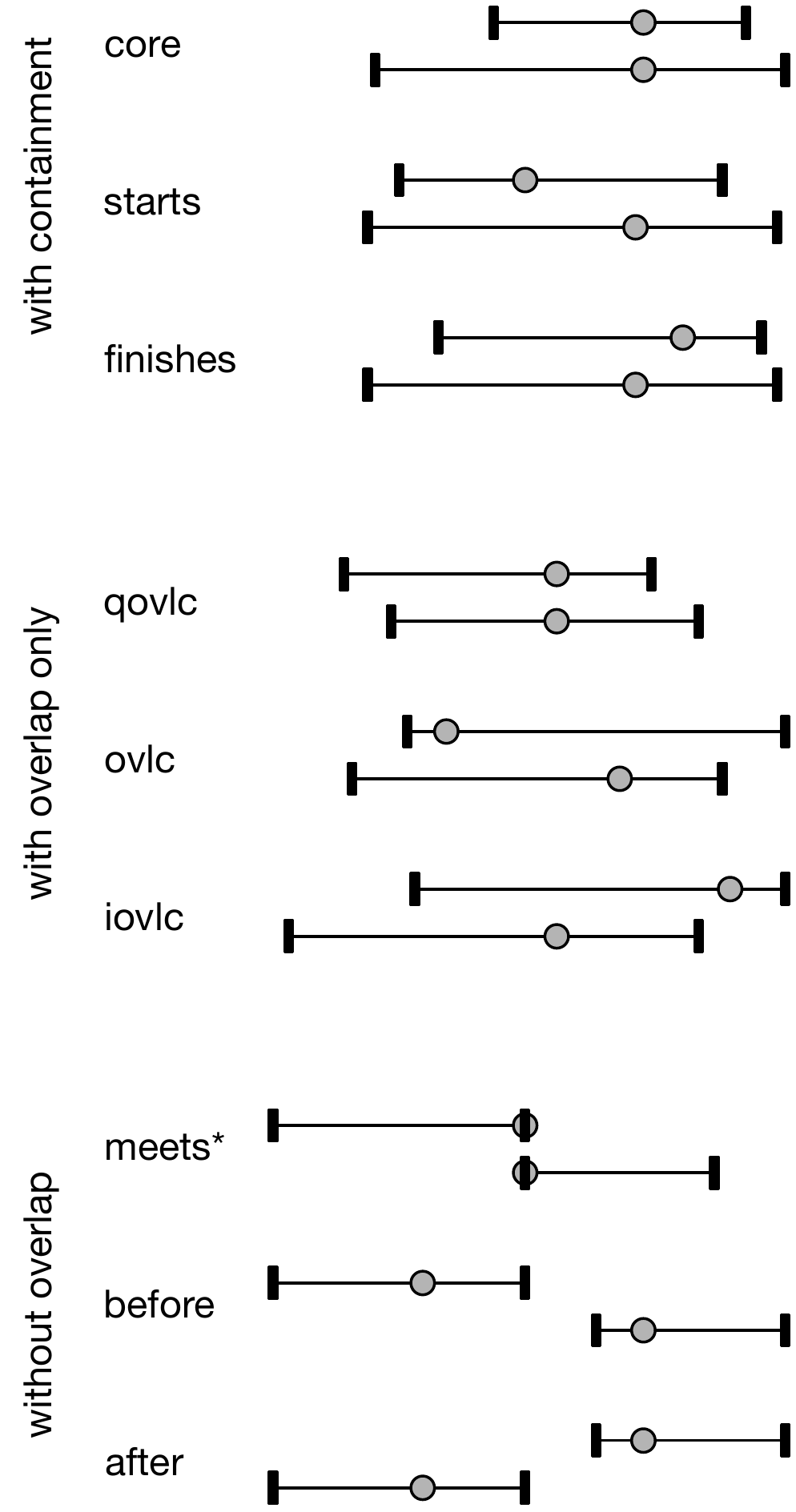}
}
\end{center}
\caption{Temporal relations based on p.o.\ relations. Left: fifteen jointly exhaustive and pairwise disjoint simple relations from two ordering relations of potential causation ($c$) and temporal containment ($t$). Here, $i \sqand t \circ j$ is an abbreviation for $[i \sqand t \ovl j] \land \lnot[j \sqand t \partof i] \land \lnot[i \sqand t \partof j]$. Right: visualized examples for orderings for potential causation (context $c$, circle point symbol) and interval containment ($t$, lines between boxes).}\figlab{intervals}
\end{figure}
Not depicted in  \figref{intervals} on the right are the inverses (\emph{icore}, \emph{istarts}, and \emph{ifinishes}) and causation variants for intervals of the same extension in $t$ (\emph{qcore}, \emph{qstarts}, and \emph{qfinishes}). Of these 15 relations 9 are transitive: \emph{core}, \emph{starts}, \emph{finishes}, \emph{icore}, \emph{istarts}, \emph{ifinishes}, \emph{qcore}, \emph{qstarts}, and \emph{qfinishes}. For the three relations with overlap (right, middle) we could further distinguish whether any starts/finishes parts of an overlapping interval are contained or not, which would produce five instead of three relations.\footnote{The notion of \emph{meets*} produced by this system of relations differs considerably from, in particular, that of \citet{allen1983maintaining} both in meaning and role for the system. 
Note, moreover, that the analogous semantics handle two different relations $c$ and $t$ as independent. That is, we arrive at 2D coordinates for time intervals as in the diagrammatic approach of \citet{kulpa2001diagrammatic} with the axis for $t$ representing the length of the intervals and the axis for $c$ representing the time line.} 
One can think about the point-based visualization for $c$ as indicating a temporal ``point of no return,'' a time point after which a certain process, which may have been going on for some time, such as the trolley approaching the people on the main track, can no longer be stopped. This notion is highly relevant in ethical problems. For example, if the agent thinks too long about the problem while standing at the lever, the trolley's passing is the point of no return. In an ethically unambiguous situation, where the side track is clear, we would consider the agent complicit in any injuries caused by inaction. Similarly, we can consider, e.g., global warming actions that target a date after the projected point of no return, as a choice of inaction. 

We can then describe the action verb sentence semantics in terms of situation or event parameters \citep{Schm2020Multi} $e_n, e_{n+1}$, where $e_n$ is the current event, $e_{n+1}$ is the newly introduced event, and the aspect $\alpha\tau$ is one of the relations in \figref{intervals}, e.g.: an intransitive verb with a prepositional adjunct \eqref{iaverb} as in ``a trolley is moving down a track'' or a transitive verb \eqref{tverb} as in ``the trolley kills a person.''
\begin{align}
&\begin{aligned}
  \angled{subj}  \angled{ver&b_i^{\sigma_v}(a,t)} \angled{pred^{\sigma_p}} \angled{obj^p} \mapsto   \\ 
  \exists e_{n+1}:\ &\alpha\tau(e_n, e_{n+1}) \land \ldots \land 
  subj(e_{n+1},\angled{subj}) \land 
  \angled{dim_v}^{\sigma_v}[e_n, e_{n+1}]\land \mbox{} \\
  &\angled{dim_p}^{\sigma_p}[e_n, e_{n+1}]\land 
  obj^p(e_{n+1},\angled{obj^p}) 
\end{aligned} \label{iaverb}\\
&\begin{aligned}
  \angled{subj}  \angled{ver&b_t^\sigma(a,t)} \angled{obj} \mapsto  \\ 
   \exists e_{n+1}:\ &\alpha\tau(e_n, e_{n+1}) \land \ldots \land 
  subj(e_{n+1},\angled{subj}) \land \mbox{} \\ 
  &
  obj(e_{n+1},\angled{obj}) \land 
  \angled{dim_v}^{\sigma_v}[e_n, e_{n+1}].
\end{aligned}\label{tverb}
\end{align}
For tho former example \eqref{iaverb}, the aspect $\alpha\tau(e_n, e_{n+1})$ of ``is moving'' resolves to the containment relation: the event of the trolley moving ($e_{n+1}$) contains the current situation ($e_{n}$): ($e_{n} \sqand t \ppartof e_{n+1}$). \emph{Move} does not have any additional dimensional semantics ($\angled{dim_v}^{\sigma_v}[e_n, e_{n+1}]$), like, e.g., \emph{enter} or \emph{climb}, but there is a prepositional adjunct ``down the track,'' specifying that the next situation is \emph{downwards} (on the track): $\angled{dim_p}^{\sigma_p}[e_n, e_{n+1}]$ resolves to ${down}[e_n, e_{n+1}]$.

\paragraph{Other Dimensions}
As examples of other dimensions, the size dimension large-small illustrates that adjectives derivable from their comparative meaning \citep{bierwisch1989semantics} fit well into the same pattern as the spatial and temporal ordering mechanisms. 
The health dimension illustrates the use of distinct points in the meaning of verbs, the intransitive ``recover'', for instance, describes health increasing from a value below the average to the average, as does the transitive use of ``heal.'' A minimal state is referenced in ``kill.'' No distinct points are given in ``harm'': an exceptionally healthy individual may be harmed and still remain above average. 
The possessive space refered to in the meanings of ``give'' and ``receive'', has the subject as a given reference point from which or to which the object is moved by the action. With the dynamic PPs based on ``to'' and ``from'' respectively a corresponding goal/source of the object can be specified in the sentence.

\subsection{Scales and Hedges}\seclab{sch}
So far, we obtained orderings of objects with the ABVM, but one may ask whether we can generate distinct points on a scale, e.g., for the semantics of \emph{small} as smaller than average in a context.  Moreover, can we distinguish \emph{small} from \emph{somewhat small} or \emph{very small}? 
For the ethical domain, under what circumstances will we be horrified by a conclusion, so as to be motivated to act?

We can discuss these questions again with the above example: $\phi = !(a\&n\&!b)\&!(b\&n\&!c)\&\ldots$. 
We get coordinate values for the minimum ($|\phi\&x\&0|=0$) and maximum ($|\phi\&x\&1|=|\phi\&x|$) for a relation $x$ and can locate a representation for the mean at $|\phi\&x|/2$. Every dimension automatically comes with such a rudimentary scale. We can thus, e.g., represent death as minimal health for the semantics of ``kill,'' as suggested above, and a small car as a car of less than size $|\phi\&x|/2$. 
The average size in this case does not require a representation of a distinctly conceptualized average sized car. We can compute a representation of a small car as $< |\phi\&x|/2$. It, inter alia, depends on the co-text and context: a large car among students may be a small car in a street scene. This is a desirable property in so far as the representation of the high amount of context dependency in human natural language use is still a challenge for NLP systems. Moreover, we do not even need to assume that the brain possesses a division mechanism. Any object that is not specifically mentioned in the text with respect to a relation, appears in the depiction at the center, because, without specification, a random number of its 1-positions are removed when a random object $r$, i.e., a random sequence of bits, is queried $|\phi\&r|$. If we know about an object $a$ that it is larger than average (size context $s$), we can construct this situation in $\phi$ by generating a random vector $r$ for one-time use and position $a$ with $!(r\&s\&!a)$ in the larger-than-average half of the space, as the expression $!(r\&s\&!a)$ removes bit positions from $\phi$ that are on $s$ but outside of $a$. 

Generalizing, further positions on the scale can be constructed. For instance, $!(r_1\&r_2\&s\&!a)$ to move $a$ closer to the average ($r_1\&r_2$ has only 50\% of the 1s of $r_1$, i.e., less bits of $s$ outside $a$ are removed). In contrast, $!((r_1|r_2)\&s\&!a)$ moves $a$ further away ($r_1|r_2$ has 50\% more 1s than $r_1$). Turning around the relation as with opposites, we can locate an object also below the average, as for ``$a$ is small'' (= smaller than average): $!(a\&s\&!r)$. We thus also obtain hedges, like ``very'' and ``somewhat'' as well as semantically related relations that differ with respect to their position on the size scale, such as ``tiny'' and ``huge.''

If we consider the ethical dimension of the Trolley Problem, we can now see how the system will evaluate the two options it is given: to kill more or less people. Both options cause pain to others. If actions causing pains to others evoke an unpleasant sensory experience in the system, e.g., disgust or anger at the agent of such an action, it will see itself as more or less disgusting, whatever answer it gives. Asking the system to decide between the two options has the only effect of making the system ``feel'' bad about itself: the position it sees itself in before the decision is higher on the moral scale than the position after any of the two actions, whatever it decides. The urge in human readers is to somehow derail the trolley and remain without having to carry the guilt of either of the bad alternatives. 
But we can also see, how the average response influences the decision. It is the degree to which the own decision is evaluated as below average that influences the strength of disgust. Where the average car is small, a small car is seen as an average car, while a normal-sized car may appear as large. 

It is inevitable that climate change will ultimately enforce a different average level of comfort. Our legacy will be judged by this standard.

\section{Discussion: Grounded Ethics}\seclab{dsc}

An intelligent machine that does no harm may be the dream of future drivers that have autonomous cars at hands. The inhibition of harmful deeds as a good moral practise may have its roots in the way we represent or imagine others' internal state or, in other words, the way we empathize. A sad face may be a source of sorrow and a catalyst of caring gestures whereas discovering a wicked motivation in a friend may awaken our aggressive instincts. Empathy may lead us to different actions and, thus, it is important to characterize how our imagination can put us on the right track.

Empathy comprises a wide range of different phenomena and, therefore, it is difficult to find a consensual definition \citep{batson2009these}. However, the current literature seems keen to state that empathy is composed by two distinct, although related, concepts: cognitive empathy and affective empathy \citep{Davi1983Measu}. Cognitive empathy refers to our inferences about others' mental states or to our ability to perceive their intentions, motivations and expectations. Affective empathy concerns our ability to share an other's feelings and emotions \citep{devig2006empathic,paulus2013distinction}. As simulation theory posits, it conveys the simulation or representation of the other's emotional experience in ourselves \citep{batson2009these}. Nevertheless, affective empathy does not mean emotional contagion, the vicarious feelings of an others' emotional state. In affective empathy, we are conscious of the other representational state as distinctive from our own \citep{devig2006empathic,paulus2013distinction}.  
  
Adopting the perspective of an other requires an act of imagination and shapes the kind of affective experience they may have \citep{Enge2011Empat}. For instance, we can imagine how the other is feeling (\emph{imagine other}) or imagine how we would feel in the other's place (\emph{imagine self}). These differences may be critical when it comes to adopting prosocial behaviours. In fact, neuroimaging studies have shown that brain systems supporting memory and imagination may shape empathy \citep{Gaes2013Const}. 
Some authors include the emotions of compassion and distress in the definition of affective empathy \citep{hodges2007empathy}. Empathic concern or compassion, namely feeling pity when imagining an other's emotions, and empathic distress, meaning the feelings of anxiety and unease at an other's suffering, are reactions to emotional sharing. They promote pro-social behaviour, facilitating socially desirable actions, leading to caring behaviour and inhibiting harmful actions \citep{Dece2014The-c,decety2015empathy}. 
  A question that may arise from this debate is what dimensions of empathy or acts of imagination bolster ethical behaviour. The answer to this question may be found in pathologies that present impairments on empathy, represented by an imbalance between cognitive empathy and affective empathy dimensions, such as psychopathy and autism \citep{smith2006cognitive,smith2009empathy}.
  
\paragraph{Psychopathy}
Psychopathy is a condition characterized by anti-social behaviour, insensibility towards an other's signs of distress (e.g. sadness or fear) and incapacity to feel certain emotions such as fear, shame, guilt or remorse \citep{Amer2013Diagn,Hare2003Manua}. Psychopaths show an inability to recognize distress cues and a lack of compassion or empathic concern. These deficits lead them to inflict serious harm on others, and even go through a life of crime \citep{Blai1995A-cog}. However, their manipulative and charming behaviour point to an understanding of others' emotions and motives, although they remain affectively callous. As a consequence, psychopaths are said to detain a high \emph{cognitive} empathy and a low \emph{affective} empathy \citep{smith2006cognitive}. This means that they are capable of understanding their victims' mental state, despite their inability to experience their victims' emotions. Such imbalance between cognitive empathy and affective empathy may inspire aggressive and competitive behavior but be at failure to induce cooperation and to avoid damage in others \citep{smith2006cognitive}. A recent study showed that psychopaths are severely impaired in imagine-other perspective rather than in imagine-self perspective \citep{Dece2013A-fMR}. In other words, they are better at representing their own emotions when imagining themselves in an other's place than at simulating an other's emotions in themselves. The failure to feel certain emotions, such as fear, prevents them from representing those emotions in their emotional system, narrowing the emotional experience \citep{BirdVidi2014Self}.
 
\paragraph{Autism} 
Autism is a neurodevelopmental disorder characterized by communication and relational problems, repetitive behaviour and disturbances in empathy \citep{Amer2013Diagn}. Autistic people are said to lack cognitive empathy, that is they have trouble in understanding others' mental states, motives or intentions (although it has been recognized that this condition improves with time, for instance, they become able to pass on false believe tests). Contrary to psychopaths, they are able to feel basic emotions such as happiness, anger, sadness and fear. Remarkably, in spite of a low cognitive empathy, their affective empathy is intact and, according to some authors, even surfeits \citep{smith2009empathy}. Therefore, they are able to feel another person's emotional state, although many times they can neither explain it nor understand it. In virtue of their ability to empathize with an other's emotional state, they exhibit signs of distress at others' distress and show signs of compassion and concern at others' disturbance. This imbalance between cognitive empathy and affective empathy may produce a caring and cooperative behaviour and avoid harmful tendencies \citep{smith2006cognitive}. In the case of autism, there is not a sharp contrast between imagine-other perspective and imagine-self perspective \citep{BirdVidi2014Self}, turning people more receptive to others' distress.
  
The model of autism empathy may be suitable to adopt if we want to build a machine that does no harm. If the machine is capable to represent in itself basic emotions such us others' fear, it will restrain itself from harmful actions. A possible objection to this architecture is that a machine conceived in this kind of framework, when facing utilitarian dilemmas, such as the trolley dilemma, probably will produce a peculiar behaviour. Some authors foresaw an enhancement in utilitarian decision-making in populations with empathy impairments, such as autism and psychopathy, for different reasons \citep{patil2015trait,vyas2017derailing}. 
  
\paragraph{Grounding in Disgust}
Disgust is an emotion characterized by avoidance of situations perceived as unclean or unhealthy. When feeling disgust, individuals tend to display lower levels of aggression and to restrain themselves from committing moral wrongdoings, as these actions are faced as unclean \citep{tindell2019examining}. Therefore, it is expectable that psychopathic subjects, who are often involved in criminal acts and exhibit high levels of aggression, show diminished feelings of disgust \citep{aharoni2012can}. On the other hand, studies regarding individuals with autism, who present compassionate behavior,  showed that they are responsive to disgusting situations \citep{zalla2011moral}.

\paragraph{Grounding via Empathy}
People with autism exhibit distress at experiencing another person's suffering and show compassion, a sign that their behavior is moral \citep{goetz2010compassion,nussbaum2003upheavals}. They even engage in practical actions designed to reduce the perceived suffering \citep{attwood2015complete,smith2009empathy}.  According to some researchers, the moral behavior performed by people with autism is a result from a preserved affective empathy \citep{bollard2013psychopathy,smith2009empathy}. In contrast, people with deficits on affective empathy, such as individuals with psychopathy, show no compassion and do not engage in moral behavior \citep{blair2007empathic}.

\section{Outlook and Conclusions}\seclab{cnc}
This article illustrated how the CL language with analogous semantics can be applied to ethical decision making. We showed how concepts of temporal reasoning from the literature on decision making and planning can be characterized in CL. We, moreover, demonstrated how the positive form of adjectives can be derived from the comparative form within the ABVM, and how we can in a similar way also represent hedges.  
For illustration purposes, we used a small natural language fragment and a description of an ethical dilemma to illustrate details of the lexicon mechanism for a neuronal logical reasoner realizing the computation of analogous semantics for atomic CL expressions. 
The article focussed on the specification of adjectives and verbs, showing that we can represent distinct scales for arbitrary sensory dimensions, with context-dependent average, minimum, maximum, and a mechanism to construct intermediate points. 
The results suggest that a grounded reasoner is a key component of human working memory (WM) required for higher cognition. We conjecture, moreover, that the many benefits of grounded reasoning created sufficient evolutionary pressure to further enlarge WM capacity, but that the complexity limits of the core \#SAT reasoner presented a strict boundary, leading to the evolution of additional step-wise depth-first reasoning mechanisms, capable of representing alternatives ($\lor,\lnot$, CL0, which has a decidable reasoning procedure) and object persistence ($\exists$, CL1 or FOL).\footnote{
The CL0/CL1 processing is detailed in \cite{Schm2020Multi}. A video of the system processing the full Trolley example can be viewed at: \url{https://logical-lateration.appspot.com/video.html}. The video shows how the CL0/CL1 reasoner circumvents the limitations of the CLA reasoner starting from a clean slate in every exploration of an alternative, thus producing a sequence of images, an inner movie, for the visuospatial domain rather than a static image.} 

We compared the notion of grounding in the proposed logical cognitive system with respect to the domain of ethical reasoning in human beings, where a lack of grounding of ethics in empathy has been proposed to be a key component for psycho-social disorders leading to unethical decision making and studied in depth for public health purposes. Interpreting these results into the proposed architecture, we can model the ethical notion of \emph{evil} as grounded in the sensory dimensions of disgust and, via empathy, pain. Generalizing, we can conclude that current autonomous vehicles even with ethical logics added will, without analogous semantics, be likely to exhibit anti-social behavior similar to that of humans with impaired grounding of ethical concepts. Leveraging logics with analogous semantics could help in this regard. Given the scalability issues of a core \#SAT mechanism for empathy and the  high scalability of perceptual capacities, we would caution that large-scale AI systems would suffer from an imbalance of considerably higher perceptual and reactional capabilities than capacity for benevolence and compassion. The current global race to develop the most powerful reactive AI systems for 
centrally processing vast amounts of perceptual user data 
may come at a dire price. 
Small-scale AI, in contrast, could excel in this regard.

\section*{Acknowledgments}
The authors wish to thank the Hanse-Wissenschaftskolleg, Delmenhorst for funding and supporting this collaboration. We are grateful to the four reviewers of this paper for valuable comments.

\end{document}